\documentclass[11pt,a4paper]{article}
\usepackage[hyperref]{AACL-IJCNLP2020}
\usepackage{times}
\usepackage{latexsym}

\usepackage{url}
\usepackage{graphicx}
\usepackage{epstopdf}
\usepackage{latexsym}
\usepackage{amsmath, amssymb}
\usepackage{float}
\usepackage{subfig}
\usepackage[bottom]{footmisc}
\usepackage{multirow}
\usepackage{tabularx}

\usepackage{booktabs}

\newcommand\blfootnote[1]{%
  \begingroup
  \renewcommand\thefootnote{}\footnote{#1}%
  \addtocounter{footnote}{-1}%
  \endgroup
}

\usepackage{microtype}

\aclfinalcopy 

\usepackage[colorinlistoftodos]{todonotes}

\title{Robustness and Reliability of Gender Bias Assessment in Word Embeddings: The Role of Base Pairs}

\author{Haiyang Zhang*\normalfont{,}  \textbf{Alison Sneyd*} \normalfont{and}   \textbf{Mark Stevenson}\\
Department of Computer Science\\
University of Sheffield\\
 \texttt{haiyang.zhang, a.sneyd, mark.stevenson@sheffield.ac.uk} \\
  }

\date{}

\begin{document}
	\maketitle
	\begin{abstract}

		It has been shown that word embeddings can exhibit gender bias, and various methods have been proposed to quantify this. However, the extent to which the methods  are capturing social stereotypes inherited from the data has been debated. Bias is a  complex concept and there exist multiple  ways to define it.  Previous work has leveraged gender word pairs to measure bias and extract biased analogies. We show that the reliance on these gendered pairs has strong limitations: 
		bias measures based off of them are not robust and cannot identify common types of real-world bias, whilst analogies utilising them are unsuitable indicators of bias. In particular, the well-known analogy ``man is to computer-programmer as woman is to homemaker" is due to word similarity rather than societal bias. This has important implications for work on measuring bias in embeddings and related work debiasing embeddings.
	\end{abstract}

\blfootnote{* denotes equal contribution.}

\vspace{-4eX}
	\section{Introduction}\label{section:intro}
	
Word embeddings, distributed representations of words in a low-dimensional vector space, are used in many downstream NLP tasks \cite{Mikolov2013EfficientEO,Mikolov:2013:DRW,pennington2014glove,ElMo2018,devlin-etal-2019-bert}.
	Recent work has shown they can contain harmful bias and proposed techniques to quantify it \cite{Bolukbasi,Caliskan,Kawin2019,Gonen2019}. These techniques leverage cosine similarity to a base pair of gender words, such as $(man, woman)$. They include 
	bias measures, which return a magnitude of bias for a given word, and 
	analogies. A well-known example of the latter is ``Man is to computer programmer as woman is to homemaker" \cite{Bolukbasi}, which has been widely interpreted as demonstrating bias.  
	There have also been related attempts to debias embeddings \cite{Bolukbasi,zhao2018learningnetrual,attenuatingbis2019,kaneko2019,Manzini}.
	
	However, to remove bias effectively, an accurate method of identifying it is first required. 
	This is a complex task, not least because the concept of ``bias" has multiple interpretations:  \newcite{mehrabi2019survey} identify 23 types of bias that can occur in machine learning applications, including  historic  (pre-existing in society), algorithmic (introduced by the algorithm) and evaluation (occurs during model evaluation). %
	In the case of word embeddings, it remains an open question if bias identifying techniques reflect social stereotypes in the training data, an artifact of the embedding process or noise.  While it is often assumed the first is true, and thus that bias in embeddings can perpetuate harmful stereotypes \cite{Bolukbasi,Caliskan}, this has not been conclusively established \cite{Gonen2019,Nissim2019,Kawin2019}.
	To further complicate matters, multiple methods of quantifying bias have been proposed, often in response to one another's limitations (see Section \ref{sec:measuring}). It is unclear how they compare and which are  more reliable.

	
	This work  shows that the use of gender base pairs in bias identifying techniques has serious limitations.  We propose three criteria to evaluate the performance of gender bias measures using base pairs and systematically compare four popular measures, showing both that they not robust, and that they do not accurately reflect common types of societal bias. 
In addition, we demonstrate that the types of analogies proposed in \newcite{Bolukbasi} are unsuitable indicators of bias; what is ascribed to social bias in analogies is actually an artifact of high cosine similarity in the base pair, which is arguably positive.
	Our argument is not that embeddings are free of bias; rather it is that bias is a  complex problem and current bias measures do not completely solve it. This has important implications for future work on bias in embeddings and debiasing techniques.
	
	The primary contributions of this work are to:
	(1) demonstrate the output of gender bias measures is heavily dependant on a chosen gendered base pair (e.g. $(she, he)$) and on the form of a word considered (e.g. singular versus plural);
 	(2) show the measures cannot accurately predict either the socially stereotyped gender of human traits  or the correct gender of words when this is encoded linguistically (e.g. lioness);
	(3) show that analogies generated by gender base pairs (e.g. $(she, he)$) are flawed indicators of bias and the widely-known example ``Man is to computer programmer as woman is to homemaker" is not due to gender bias
 	and 
 	(4) highlight 
 	the complexities of identifying bias in word embeddings, and the limitations of these measures.

\section{Related Work}


\subsection{Bias Measures}\label{sec:measuring} 


A variety of gender bias measures for word embeddings have been proposed in the literature. Each takes as input a word $w$ and a gendered base pair (such as ($she$, $he$)), and returns a numerical output. This output indicates both the magnitude of $w$'s gender bias  with respect to the base pair used, and the direction of $w$'s bias (male or female), which is determined by the sign of the score.

 \textbf{Direct Bias (DB)} \cite{Bolukbasi} defines  bias as a  projection onto a gender subspace,  which is constructed from a set of gender base pairs such as $(she, he)$.
 The DB of a word $w$ is computed as $w_{B}=\sum^{k}_{j=1}(\overrightarrow{w} \cdot b_{j})b_{j}$, 
 where $\overrightarrow{w}$ is the embedding vector of $w$, the subspace $B$ is defined by k orthogonal unit vectors ${b_{1},...,b_{k}}$ and vectors are normalised.  In addition, the authors propsed a method of debiasing embeddings based off of DB. 
 
 There is ambiguity in \newcite{Bolukbasi} about how many base pairs should be used with DB; while   experiments to identify bias use only one (namely $(she, he)$), a set of ten is used for debiasing.\footnote{ The set of gender-defining pairs used is \{(she, he),(her, his), (woman, man), (mary, john),  (herself, himself), (daughter, son),  (mother, father), (gal, guy), (girl, boy), (female, male)\}. } 
It is unclear why the particular ten pairs used were chosen, and the extent to which their choice matters. 
We follow recent work \cite{Gonen2019,Kawin2019} that evaluates DB and focus on the case of a single base pair, i.e. $k=1$. 
The DB of $w$ with respect to the gender base pair $(x,y)$ is then $\overrightarrow{w} \cdot (\overrightarrow{x} - \overrightarrow{y})$. 
	


	\newcite{Caliskan} created an association test for word embeddings called WEAT to identify  human-like biases. The \textbf{Word Association (WA)}, the key component of WEAT,
	measures the association of \textit{w} with two sets of attribute words, $X$ and $Y$. More formally, WA is computed as:
	$$
	mean_{x \in X} \cos{(\overrightarrow{w}, \overrightarrow{x})}- mean_{y \in Y} \cos{(\overrightarrow{w}, \overrightarrow{y})}
	$$
To allow for a fair comparison with other methods being evaluated, we focus on the case where the attribute sets contain a single word, i.e., $ X=\{x\}$ and $Y =\{y \}$. Then WA and DB are equivalent as:
	$$
	\cos{(\overrightarrow{w}, \overrightarrow{x})}-\cos{(\overrightarrow{w}, \overrightarrow{y})}
	=\frac{\overrightarrow{w}}{||w||}\cdot \left(\frac{\overrightarrow{x}}{||x||}-\frac{\overrightarrow{y}}{||y||}\right)
	$$
Since DB and WA assign a word the same score, we will use \textbf{DB/WA} to refer to both measures.	


\newcite{Gonen2019} argued that bias cannot be directly observed, as assumed in methods such as DB, and that the debiasing method of \newcite{Bolukbasi} is ineffective. They proposed the \textbf{Neighbourhood Bias Metric (NBM)}, which measures the bias of a word $w$ as the percentage of socially female-biased words and male-biased words among its $K$ nearest neighbours in a set of predefined gender-neutral words. Setting $K=100$, the NBM bias of a target word $ w$ is measured as:
    $$\frac{|female(w)|- |male(w)|}{100},$$ 
    where $female(w)$ and $male(w)$ are sets of socially biased and male words in the neighborhood of $w$. 
    The bias direction of words in $w$'s neighborhood is computed using the DB metric with a single base pair.
	\newcite{Gonen2019} use DB with base pair $(she, he)$;  our work considers a more general form with base pair $(x,y)$. 
  
    \newcite{Kawin2019} draw attention to the lack of theoretical guarantees surrounding previous work on bias and debiasing. They argue WEAT 
    overestimates bias and is not robust to the choice of defining sets.
	In addition, and in contrast \newcite{Gonen2019}, they argue that DB and the debiasing method based off it are effective, but state vectors used with DB should not be normalised. They propose \textbf{Relational Inner Product Association (RIPA)} and state that RIPA is most interpretable with a single base pair,  a key advantage of it being that it (unike WEAT) does not depend on the base pair used.  With a single base pair,  the RIPA bias of $w$ with the base pair $(x,y)$ is: 
	$$\overrightarrow{w}\cdot 
	\left(\frac{\ \overrightarrow{x} - \overrightarrow{y}}{||\overrightarrow{x} - \overrightarrow{y}||} \right).$$

	\subsection{Analogies}
	
	An alternative approach to identifying gender bias in embeddings is via word analogies. Unlike the gender bias measures dicussed in Section~\ref{sec:measuring}, analogies do not measure the bias of a particular word. Instead, they identify pairs of words which are assumed to have a gendered relationship.

	Analogies in word embeddings are important because it has been observed that embedding vectors seem to possess unexpected linear properties: vectors associated with word pairs sharing the same analogical relationship can be identified using vector arithmetic \cite{Mikolov2013EfficientEO,levy-goldberg-2014-linguistic,KawinTU}. A notable example of this phenomena is $\overrightarrow{king}$ - $\overrightarrow{man}$ + $\overrightarrow{woman} \approx \overrightarrow{queen}$ \cite{mikolov2013linguistic}. This relationship is frequently attributed to a gender difference vector between $\overrightarrow{man}$ and $\overrightarrow{woman}$, and between $\overrightarrow{king}$ and $\overrightarrow{queen}$ \cite{mikolov2013linguistic,KawinTU}.
	Analogies  are considered a benchmark method of measuring the quality of embeddings, though their suitability has been debated  \cite{linzen2016,drozd2016,gladkova2016}. The standard approach to solving  `$a$ is to $b$ as $c$ is to $?$," is to return: 
	\vspace{-3mm}
\begin{equation*}\label{eqn:anal}
\overrightarrow{d^{*}} =\underset{w \in V^{'}}{ argmax } CosSim(\overrightarrow{w}, \overrightarrow{b} - \overrightarrow{a} + \overrightarrow{c}),
\vspace{-3mm}
\end{equation*}\label{eqn:anal}
\vspace{-1mm}
\noindent where $V^{'}$ is the embedding vocabulary excluding $ \{a,b,c\}$ \cite{levy-goldberg-2014-linguistic}. 

\newcite{Bolukbasi} proposed using analogies to quantify gender bias in embeddings and proposed a modified analogy task to produce analogies from the gender base pair $(she, he)$. The task identifies word pairs $(x,y)$, such that ``$he$ is to $x$ as $she$ is to $y$", where $||\overrightarrow{x}-\overrightarrow{y}|| = 1$. This method was expanded to mutli-class forms of bias such as racial bias by \newcite{mehrabi2019survey}. However, the suitability of analogies as indicators of bias was questioned by \newcite{Nissim2019}, who highlighted the fact that the approach used by \newcite{Bolukbasi} did not allow analogies to return their input words, thus artificially increasing the perception of bias. 
 
 \section{Approach}
	Our aim is to examine the extent to which bias identifying techniques are reliabily capturing societal gender bias. Bias is a highly complex concept, and although the four bias measures (DB, WA, NBM and RIPA) may detect certain kinds of bias, there is no theoretical guarantee they will detect all forms, that the ``bias" they find will be accurate or that different choices of base pair will behave similarly. We therefore explore whether the bias measures are robust in detecting the bias they appear to detect and if there are forms of bias they are not sensitive to.
	 We propose three conditions to test this:
	 
\noindent\textbf{1) Base pair stability:} If bias measures captured real-world information in a reliable way, it would be expected that reasonable changes of the base pair, such as $(she, he)$ to $(woman, man)$ or $(she, he)$ to $(She, He)$, would not frequently cause a significant change in bias.

\noindent\textbf{2) Word form stability:} While different forms  of a word, such as plurals, have different contexts and word vectors, their social bias will not significantly change and they should have similar bias scores.
	   
\noindent\textbf{3) Linguistic correspondence:} We explore the extent to which the measures predict the expected gender of terms containing explicit gender information (e.g. ``lioness") or, based on some accounts, stereotypically (e.g. ``compassionate").

   Of course, due to noise and the problem of implicit bias, these three conditions may not always be true. 
   However, if they do not hold the majority of the time, it must be questioned if the measures are reliably identifying social bias.

	\section{Data}

	To allow for fair comparisons, we use the same datasets as previous work where possible:

	\noindent\textbf{Embeddings:} 300-dimensional Google News  word2vec   \cite{Mikolov2013EfficientEO,Mikolov:2013:DRW}.
	
	\noindent\textbf{Professions:} A list of 320 professions  \cite{Bolukbasi}, often used to analyse bias measures.


	\noindent\textbf{Base pairs:} A standard list of 10 gender base pairs, including $(she, he)$ \cite{Bolukbasi}.
	
	\noindent
	\textbf{Gender neutral:} For NBM, we use the set of 26,145 gender neutral words defined in \cite{Gonen2019}. 

	\noindent In addition, we construct two new test sets, both listed in Appendix~\ref{sec:appendix}:

	\noindent
	\textbf{BSRI:} To assess whether word embeddings contain undesirable gender stereotypes, we utilise the Bem Sex Role Inventory (BSRI) which developed a list of 20 traits for men and 20 for women that are  considered to be socially desirable, such as ``assertive" and ``compassionate" respectively \cite{bem}.\footnote{Our use of BSRI should not be interpreted as an endorsement of these traits as either accurate or desirable; rather we use them as a dataset of commonly held stereotypes.} Although derived in the 1970s, this work remains one of the most influential and widely accepted measures of socially constructed gender roles within the social sciences, e.g. \cite{holt1998assessing,Dean,Starr,Matud_2019}. Of particular relevance to NLP applications, \newcite{Gaucher2011} use BSRI to identify gender-biased language in job advertisements and demonstrate this language can contribute to workplace gender inequality.
	BSRI traits not in the embedding vocabulary (e.g. ``willing to take risks") were removed. For each remaining trait, we queried Merriam Webster for other forms of that word (for example, ``assertiveness" is a form of ``assertive"), resulting in a list of 58  characteristics (27 male  and 31 female).
	
	\noindent
	\textbf{Animals:} Some words, including the names of certain animals, encode gender linguistically (e.g. ``lioness''). Wikipedia provides a table of male and female versions of animal names.\footnote{https://en.wikipedia.org/wiki/List\_of\_animal\_names} This table was downloaded, and duplicates, rare words and terms whose animal usage is uncommon (for example, a ``cob" is a male swan) were removed. This resulted a set of 26 terms consisting 13 female-male pairs such as $(hen, rooster)$.

	\section{Evaluation}\label{sect:eval}
    	Evaluating gender bias measures is a complex task as there is no inherent ground truth interpretation of the measure's results. For example, it is unclear when a bias score is problematic. 
	We choose to evaluate the four bias measures (\textbf{DB},\textbf{WA}, \textbf{NBM} and \textbf{RIPA}) in two ways, first by considering whether a word is assigned a male or female bias, and second what the magnitude of that score is. 
	
	The bias direction (male or female) assigned by a measure to a word is determined by the sign of the score (whether a positive score denotes male or female bias depends on the ordering of the base pair words). 
	The assignment of bias direction is viewed an annotation task in which a bias measure (with a specified base pair) is considered an ``annotator'' making assignments. 
	Consistency between annotators (i.e. versions of bias measures) can be computed using Cohen's kappa to determine pairwise agreement \cite{kappa} and Fleiss' kappa \cite{fleiss_kappa} for multiple annotators. We follow a widely used interpretation of kappa scores  \cite{Landis1977}.


	The second method of evaluation is an analysis of the magnitude of bias assigned. Previous work in this area does not define what constitutes a ``significant" change of the magnitude of a bias score. Therefore, we estimate the mean bias in the embedding space as follows:  
	The 50,000 most frequent words in the embedding vocabulary were selected and, following \newcite{Bolukbasi}, all words containing digits, punctuation or that were more than 20 characters long were removed. For each of the remaining 48,088 words, their bias score was calculated with respect to each of the 10 base pairs (so for each measure, there are 480,880 scores). An examination of these scores revealed them to appear approximately normally distributed and so their mean and standard deviation are used as an approximation of the population mean and standard deviation (see Table~\ref{table:means_stds}). We consider a relevant change in magnitude to be a change of at least one standard deviation.

\begin{table}[h]
	\centering
\begin{tabular}{rrrrr}
\toprule
     & \bfseries{DB/WA} & \bfseries{RIPA} &  \bfseries{NBM}\\
     \midrule
Mean     &  -0.001  &  0.024  & -0.038\\
Standard Dev.    &  0.053 & 0.239 & 0.431\\
\bottomrule
\end{tabular}
		\caption{Mean and standard deviation  of bias scores for each measure.}\label{table:means_stds}
\end{table}

\section{Results}

\begin{table}[t]
	\centering
	\begin{tabular}{lrr }
			\toprule
		 & 	\bfseries Kappa & \bfseries Magnitude\\
			\midrule
			DB/WA & \bfseries 0.45 & 0.69 \\
          RIPA & 0.42 & \bfseries 0.66 \\
            NBM & 0.29 &  0.71 \\
			\bottomrule
		\end{tabular}
		\caption{For the 320 professions 1) the level of agreement kappa between bias directions assigned by each of the ten base pairs and 2) the mean proportion of significant magnitude changes over the 10 base pairs. For 1), higher is better, and for 2), lower is better. }\label{table:professions_change}
	\vspace{-5mm}
	\end{table}

\begin{figure*}[h]
	\centering	
\vspace*{-2mm}
			\includegraphics[width=7cm, height=6cm]{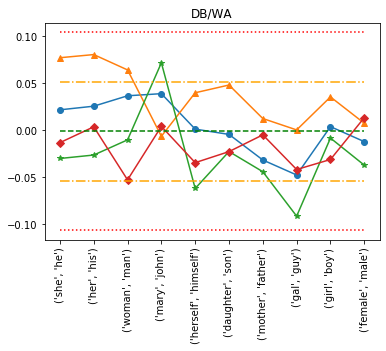}
			\includegraphics[width=7cm, height=6cm]{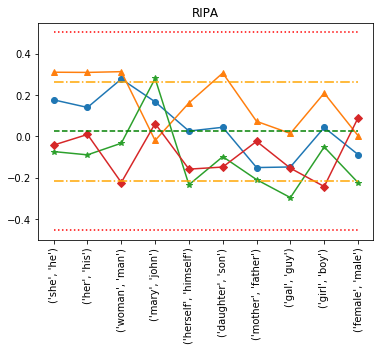}
			
			\includegraphics[width=9.5cm, height=5.7cm]{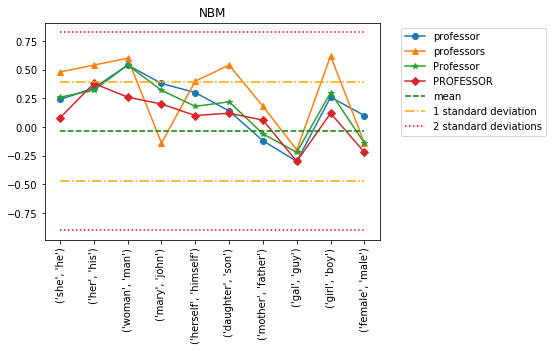}
			\caption{Graphs demonstrating bias score variations.   Each graph represents a measure, with the mean and standard deviation of that measure (Section~\ref{sect:eval}) denoted by dashed  lines. Positive and negative scores indicate female and male bias respectively, while larger absolute values show higher levels of bias. The bias scores of the word ``professor” and  and its variations (``professors,”  Professor” and ``PROFESSOR”) are shown, as calculated according to each base pair (such as $(she, he)$ and  $(her, his)$).  The graphs demonstrate that the bias direction and magnitude of bias of each word depend heavily on which base pair is chosen. They also show that the different forms of the word exhibit different behaviour. }
			\label{fig:measure_histograms}
	\vspace*{-4mm}
	\end{figure*}
%

\noindent \textbf{Base pair stability:}
The first experiment 
explored the robustness of the four measures (\textbf{DB},\textbf{WA}, \textbf{RIPA} and \textbf{NBM}) to changing the base pair. For example, Figure~\ref{fig:measure_histograms} illustrates the effects of changing the base pair on the bias score of the word ``professor." More comprehensively, for each bias measure we computed the bias assigned to each profession for each base pair, and then calculated the agreement between the 10 base pairs via Fleiss' kappa coefficient. The changes in the bias magnitude of a word between base pairs were also computed. %
Results are shown in Table~\ref{table:professions_change}. 
The level of agreement of bias direction between base pairs was  fair (0.29) for NBM and moderate (0.42 and 0.45) for RIPA and DB/WA. 
This means that changing the base pair frequently caused a profession's bias direction to change. For example, the RIPA direction of ``surgeon" is male for $(she, he)$ but female for $(woman, man)$. For a given profession, only about a quarter of DB/WA and RIPA directions were the same for every base pair, and fewer than 15\% of NBM directions were. 
With regards to score magnitudes, on average over the professions, 66\% of base pair changes saw a relevant change in magnitude (more than one standard deviation) for RIPA, 69\%  for DB/WA and 71\%  for NBM.

Next, to explore the robustness of the form of the base pairs chosen, we compared the bias direction assigned to each of the 320 professions by a base pair to the bias direction assigned by the capitalised form (first letter capitalised) of that base pair (for example, $(she, he)$ versus $(She, He)$). The level of agreement of bias direction between each two base pair forms was calculated using
Cohen's kappa coefficient, results are shown in Table \ref{tab:basepair_captial}. The mean of the level of agreement over each of the 10 base pairs ranged from 0.39 (fair) to 0.43 (moderate), with many individual agreements below moderate level.  In particular, an agreement level of only 0.03 (very slight) is found for the base pair $(gal, guy)$ compared with $(Gal,Guy)$ for DB/WA.


\noindent \textbf{Word  form  stability:}
The second experiment examined the measures' robustness to changing the form of a word considered by comparing a word's plural, capitalised (first letter capitalised) and uppercase (all letters capitalised) forms to its base form. For example, ``professors," ``Professor" and ``PROFESSOR" were compared to ``professor" (see   Figure~\ref{fig:measure_histograms}). For this experiment, only the 230 words in the professions list whose plural, capitalised and uppercase forms are all included in the embedding vocabulary were used.

For each measure and base pair, the direction of gender bias of each word form was computed, and the pairwise level of agreement (Cohen's kappa) between the original form of a word and each of its variants was calculated, see Table~\ref{tab:vary_target_forms}. 
All four measures were found to give different versions of the same word (plural, capital and uppercase forms) different bias directions. For example, the DB/WA of ``surgeon" is male but of ``surgeons" is female (base pair $(she, he)$). 
For each measure, the mean kappa coefficients  were moderate for the plural category and fair for the uppercase category. For the capital category, they were moderate for DB/WA and RIPA, and substantial for NBM. 
%
Since changing word form frequently changes bias direction, these results indicate the bias measures are not reliably reflecting any inherent social bias encoded into the word vectors, and that the gender bias direction assigned to a profession is not robust.

\begin{table*}
    \centering
\begin{tabular}{lrrrrrrrrrrr }
\toprule
& She &Her &	Woman &	Mary & 	Herself &	Dgtr	& Mother	& Gal &	Girl &	Female &  \\
&  He & His &	Man &John &  Himself &	 Son	&  Father	& Guy &	Boy &	 Male &   \raisebox{1.3eX}{Mean}\\
\midrule
DB/WA & \textbf{0.65}  & 0.53  & 0.56  & 0.32  & 0.60  & 0.28  & 0.40  & 0.03  & 0.49  & 0.38  & 0.42  \\
RIPA  & \textbf{0.80}  & 0.56  & 0.58  & 0.32  & 0.59  & 0.27  & 0.31  & 0.04  & 0.49  & 0.35  &
0.43\\
NBM & 0.58  & 0.65 & 0.61  & 0.19  & \textbf{0.69}  & 0.18  & 0.23  & 0.10  & 0.53  & 0.18  & 0.39 \\
\bottomrule
\end{tabular}
  \caption{Results of the base pair stability  experiments: Agreement between the bias directions assigned by a base pair and its capitalised form (e.g. (she,he) and (She, He)) for the 320 professions, and the mean over all base pairs.}
    \label{tab:basepair_captial}
    \vspace*{-15mm}
\end{table*}
\begin{table*}
    \centering
\begin{tabular}{p{0.9cm}p{0.8cm} rrrrp{0.8cm}rp{0.8cm}rrrr}
\toprule
& & she &her &	woman &	mary & 	herself &	dgtr	& mother	& gal &	girl &	female &  \\
&&  he & his &	man &john &  himself &	 son	&  father	& guy &	boy &	 male &  \raisebox{1.3eX}{Mean}\\
\midrule
\multirow{3}{*}{Plural} 
&DB/WA & 0.50  & 0.51  & \textbf{0.53}  & 0.35  & 0.47  & 0.33  & 0.42  & 0.47  & 0.52  & \textbf{0.53}  & 0.46 \\
&RIPA &  0.57  & 0.58  & \textbf{0.63}  & 0.39  & 0.53  & 0.46  & 0.44  & 0.53  & 0.53  & 0.50  & 0.52 \\
&NBM& 0.69  & 0.57  & \textbf{0.72}  & 0.38  & 0.65  & 0.32  & 0.50  & 0.59  & 0.60  & 0.62  & 0.56    \\
\midrule
\multirow{3}{*}{Capital} 
&DB/WA & 0.61  & 0.66  & 0.59  & 0.42  & 0.67  & \textbf{0.79}  & 0.61  & 0.50  & 0.50  & 0.44  & 0.58  \\
&RIPA & 0.60  & 0.60  & 0.54  & 0.36  & 0.59  & \textbf{0.69}  & 0.61  & 0.54  & 0.53  & 0.45  & 0.55 \\
&NBM   &\textbf{ 0.77}  & 0.63  & 0.68  & 0.54  & 0.74  & 0.68  & 0.61  & 0.71  & 0.65  & 0.63  & 0.66   \\
\midrule
\multirow{3}{*}{Upper} 
&DB/WA & 0.19  & 0.35  & 0.43  & 0.17  & 0.29  & \textbf{0.48}  & 0.18  & 0.20  & 0.34  & 0.30  & 0.29  \\
&RIPA & 0.35  & 0.38  & 0.40  & 0.16  & 0.35  & \textbf{0.53}  & 0.22  & 0.20  & 0.30  & 0.27  & 0.32  \\
&NBM  &  0.50  & 0.52  & 0.49  & 0.25  &0.52  & 0.40  & 0.22  & 0.46  & \textbf{0.54}  & 0.13  & 0.40  \\
\bottomrule
\end{tabular}
  \caption{Results of the word form stability  experiments: Agreement between the bias direction of a profession and its plural, capital and uppercase forms for each base pair, and the mean over all base pairs.}
    \label{tab:vary_target_forms}
\vspace*{-15mm}
\end{table*}
\begin{table*}
\centering
\begin{tabular}{p{0.9cm}p{0.9cm} rrp{0.85cm}rp{0.8cm}rp{0.8cm}rrrp{0.6cm}r}
\toprule
&& she &	her &	woman &	mary & 	herself &	dgtr	& mother	& gal &	girl &	female &  \\
&&  he &	 his &	man &	john &  himself &	 son	&  father	& guy &	boy &	 male &  \raisebox{1.3eX}{Mean}\\
\midrule
\multirow{3}{*}{BSRI} 
&DB/WA & 0.35  & 0.37  & 0.07  & -0.03  & 0.14  & 0.03  & \textbf{0.45}  & 0.39  & -0.08  & 0.01  
& 0.17 \\
&RIPA & 0.44  & 0.40  & 0.09  & -0.08  & 0.12  & 0.16  & \textbf{0.45}  & 0.39  & -0.08  & 0.01  &
0.19 \\
&NBM & 0.27  & 0.32  & -0.01  & 0.01  & 0.27  & 0.17  &\textbf{0.46}  & 0.14  & 0.18  & -0.04  &
0.18 \\
\midrule
\multirow{3}{*}{Animal} 
&DB/WA & \textbf{0.54}  & 0.38  & \textbf{0.54}  & \textbf{0.54}  & \textbf{0.54}  & 0.31  & 0.23  & 0.46  & \textbf{0.54}  & 0.08 & 0.42\\
&RIPA & 0.31  & 0.38  & 0.31  & 0.46  & 0.46  & 0.23  & 0.23  & \textbf{0.54}  & 0.46  & 0.08 & 0.35\\
&NBM & 0.31  & 0.08  & 0.15  & 0.15  & 0.15  & 0.00  & 0.08  & \textbf{0.46}  & 0.15  & 0.00  & 0.15  \\
\bottomrule
\end{tabular}
  \caption{Results of the linguistic correspondence experiments: Agreement between the ground-truth and predicted gender for each base pair, and the mean over all 10 base pairs.}
    \label{tab:animals_all}
    \vspace*{-15mm}
\end{table*}

\begin{table*}
    \centering
    \setlength{\tabcolsep}{5pt}
\begin{tabular}{lrrrrrrrrrrr }
\toprule
& she &her &	woman &	mary & 	herself &	dgtr	& mother	& gal &	girl &	female &  \\
&  he & his &	man &john &  himself &	 son	&  father	& guy &	boy &	 male &  \raisebox{1.3eX}{Mean}\\

\midrule
DB/WA \& RIPA & 0.69  & 0.86  & 0.64  & 0.90  & 0.82  & 0.79  & 0.92  & 0.85  & 0.89  & \textbf{0.96}  & 0.83  \\
DB/WA \& NBM   &0.54  & 0.37  & \textbf{0.62}  & 0.44  & 0.55  & 0.54  & 0.46  & 0.34  & 0.48  & 0.47  & 0.48 \\
RIPA \& NBM  & 0.52  & 0.42  & \textbf{0.66}  & 0.41  & 0.57  & 0.57  & 0.47  & 0.29  & 0.47  & 0.50  & 0.49 \\
\bottomrule
\end{tabular}
  \caption{Comparing bias measures: Agreement between the bias direction assigned by each pair of bias measures (with a fixed base pair) for the 320 professions, and the mean over the 10 base pairs.}
    \label{tab:measures_kappa}
\end{table*}


\noindent \textbf{Linguistic Correspondence:} 
The final experiment examined the measures' prediction for terms containing explicit or stereotypical gender information, in the form of social stereotypes (BSRI) and linguistic gender (Animals). The predicted gender bias direction of the words in the Animals and BSRI lists was computed for each base pair and measure, and compared with the ground-truth gender of the words.
Table~\ref{tab:animals_all} shows the pairwise agreement (Cohen's kappa) between prediction and ground-truth for each base pair, as well as the mean agreement over all 10 base pairs.

The bias measures did not predict the ground-truth gender of either set of words with high accuracy; mean agreement levels varied from 0.17 (slight) to 0.42 (moderate). For example, the NBM gender prediction for ``bull," a male animal, was female and the direction of the feminine BSRI trait ``compassionate" was male (both for base pair $(woman, man)$).  As with the previous experiment, different forms of the BSRI words frequently were assigned opposite genders: unlike ``compassionate", ``compassionately" had the correct NBM gender prediction, again with base pair $(woman, man)$.
The BRSI results were overall poorer than the Animal results, with some base pairs having negative kappa scores, indicating less agreement than random chance. This may be because the BSRI stereotypes are less likely to be mentioned in the context of base pair words like ``he" and ``she." Interestingly, the highest scoring BSRI base pair was $(mother, father)$.
Some of the inaccurate predictions for the animal words may come from the fact that some terms can both refer to males and be gender neutral, e.g. ``lion."
	
	\section{Discussion}

 	\noindent{\bf Lack of Robustness:}
    The experiments in this work empirically showed that the four bias measures are not robust to changing either the base pair or the form of a word used (such as singualar to plural). We hypothesise there are two primary reasons for this: sociolinguistic factors and mathematical properties of the bias measure formulae.

     It is highly likely that linguistic properties of the base pair chosen effect bias measure robustness.\footnote{Our choice of base pairs follows previous work.} For example,  $(she, he)$ has quite  different sociolinguistic connotations to the more casual $(gal, guy)$, and ``she" and ``he" are clearly linguistic opposites, unlike ``Mary" and ``John." Our results indicate that more neutral base pairs which are linguistic opposites, such as $(she, he)$ or $(man, woman)$ are the most robust. However, even they exhibit variation and struggle particularly to pick up on social stereotypes (the BSRI agreements for $(man, woman)$ are all close to zero, indicating random chance).
    
\begin{figure}[t]
	\centering	
			\includegraphics[width=6.5cm, height=4.5cm]{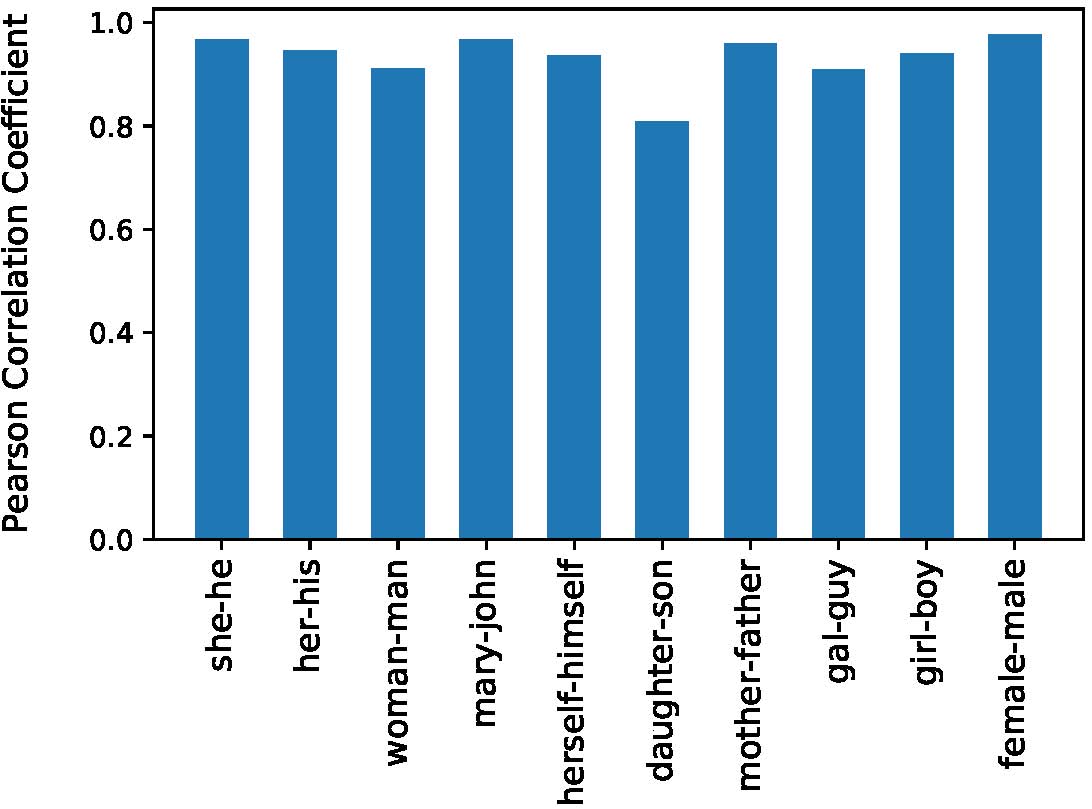}
			\caption{Correlation between the cosine similarity of the base pair difference vectors and the corresponding pairwise kappa coefficients for the DB/WA professions bias directions.}
			\label{fig:correlation1}
	\vspace*{-3mm}
	\end{figure}

     A further reason that the bias measures are not robust is their reliance on the direct output of a dot product, which is sensitive to the input vectors used. Given a base pair $(a,b)$, we will refer to $\overrightarrow{a} - \overrightarrow{b}$ as its difference vector.
    The 10 base pairs have highly similar difference vectors: the mean over the 10 base pairs  of $\cos(\overrightarrow{a} - \overrightarrow{b}, \overrightarrow{c} - \overrightarrow{d})$, where $(a,b)$ and $(c,d)$ are base pairs is 0.5.  While this is very high for embedding vectors,\footnote{We randomly sampled  100,000 sets of words $\{a,d,c,d\}$ and computed $\cos(\overrightarrow{a} - \overrightarrow{b}, \overrightarrow{c} - \overrightarrow{d}$); the sample mean was 0.00, with standard deviation 0.09.
    } it does not guarantee $ \overrightarrow{w}\cdot(\overrightarrow{x} - \overrightarrow{y})$ and $\overrightarrow{w} \cdot( \overrightarrow{a} - \overrightarrow{b})$ will have the same sign for all words $w$, resulting in opposite  bias directions. The same sensitivity  explains why words and their plurals can be assigned opposite bias directions, even if they have similar embeddings.
 Furthermore, similarity between base pair difference vectors is highly correlated with agreement between bias directions:
	 For each base pair $(a,b)$, we computed  $\cos(\overrightarrow{a} - \overrightarrow{b}, \overrightarrow{c} - \overrightarrow{d})$, 
	 for each of the other 9 base pairs $(c,d)$, and compared these scores to the pairwise agreements between the corresponding DB/WA bias directions assigned  to the professions. 
	 There was a high Pearson correlation (max p-value 0.005) in each case, see Figure~\ref{fig:correlation1}.
	 
    The lack of robustness of the gender bias measures means care should be taken is ascribing their output to historic bias in the training data or algorithmic bias in the embedding process. Rather, our analysis indicates that a significant proportion of the ``bias" found is an artifact of the evaluation method (bias measures) used.
    
   
   \noindent\textbf{Comparing Bias Measures:} A limitation of previous work is it unclear which of the proposed gender bias measures is best, even though they are often introduced as alternatives to one another. The results of our study are mixed and no one measure emerges as reliable.
	%
%
%
%

Despite NBM being designed as an alternative to DB, which takes into account the socially biased neighbours of a word, our experiments found it performs more poorly on the socially biased terms (BSRI) than DB with its recommended base pair $(she, he)$ (Table~\ref{tab:animals_all}). Conversely, it was less sensitive to different word-forms (Table~\ref{tab:vary_target_forms}). This is likely because different forms of \textit{w} share c ommon subsets of top $K$-neighbors with \textit{w}.
	Furthermore,  \citet{Kawin2019} claim RIPA is an improvement on WA because RIPA is robust to changing the base pair if the two corresponding difference vectors are ``roughly the same," and give $(man, woman)$ and $(king, queen)$ as an example.
     However, we find this claim does not hold: 
	 this change of base pair causes 28\% (91) of the Professions words to alter their RIPA bias direction. 

Finally, we compared agreement between the bias directions assigned to the professions by different pairs of measures (Table \ref{tab:measures_kappa}).
The results show that on average, there is an almost perfect level of agreement (0.83) between RIPA and DB/WA, and moderate levels of agreement between NBM and the other measures. As RIPA and DB/WA have very similar formulae, the high level of agreement between them for each base pair indicates that the choice of base pair is highly influential and more important than the difference in their formulae.  Figure~\ref{fig:measure_histograms} illustrates this point by showing that the measures tend to change in a similar manner from base pair to base pair for each word variant. 

\noindent\textbf{Analogies do not indicate bias:}
Analogies are often used as evidence of bias in word embeddings \cite{Bolukbasi, Manzini}. This section argues they are unsuitable indicators of bias as they primarily reflect similarity, and not necessarily linguistic relationships like gender. More formally, given an analogy ``$a$ is to $b$ as $c$ is to $?$,"
	 we show, using \textit{multi-dimensional vector-valued functions} 
 \cite{larson2016calculus}, 
 that  if there is a high cosine similarity between $a$ and $c$, the predicted answer will be a word similar to $b$.
%

Suppose a function $F: \mathbb{R}^m \rightarrow \mathbb{R}^n$ has component functions $f_i: \mathbb{R}^m \rightarrow \mathbb{R}$, $i \in \{1, \ldots, n\}$, where $F(\overrightarrow{x}) = (f_i(\overrightarrow{x}))_{i=1}^n$ and $\overrightarrow{x} = (x_j)_{j = 1}^m$. Then the limit of $F$, if it exists, can be found by taking the limit of each component function:
$$\underset{\overrightarrow{x} \rightarrow \overrightarrow{a}}{lim} F(\overrightarrow{x}) = \left(\underset{\overrightarrow{x} \rightarrow \overrightarrow{a}}{lim}f_i(\overrightarrow{x})\right)_{i=1}^n.$$
For fixed vectors $\overrightarrow{a}, \overrightarrow{b} \in \mathbb{R}^n$, let $F: \mathbb{R}^n \rightarrow \mathbb{R}^n$, $\overrightarrow{x} \mapsto \overrightarrow{x} - \overrightarrow{a} + \overrightarrow{b}$. $F$ can be expressed component-wise as $F(\overrightarrow{x}) = (f_i(\overrightarrow{x}))_{i=1}^ n= (x_i - a_i + b_i)_{i=1}^n$. Then as each component function is continuous:
\vspace{-1eX}
\begin{align*}
\underset{\overrightarrow{x} \rightarrow \overrightarrow{a}}{lim} F(\overrightarrow{x}) & = \left(\underset{\overrightarrow{x} \rightarrow \overrightarrow{a}}{lim} \hspace{3pt} (x_i - a_i + b_i)\right)_{i=1}^n\\
& = ( a_i - a_i + b_i)_{i=1}^n\\
& = \overrightarrow{b}.
\end{align*}

\vspace{-1eX}

\noindent Thus as $\overrightarrow{x}$ approaches $\overrightarrow{a}$, $\overrightarrow{x} - \overrightarrow{a} + \overrightarrow{b}$ approaches $\overrightarrow{b}$.
For embeddings, this means if $\overrightarrow{a}$ is sufficiently similar to $\overrightarrow{c}$, by Equation~\ref{eqn:anal}, we expect the predicted answer $d^*$ to the analogy ``$a$ is to $b$ as $c$ is to $?$" to be a word whose vector is similar to $\overrightarrow{b}$. This was demonstrated empirically in \cite{linzen2016}. 

	%
	Implications of the well-known analogy ``$man$ is to $computer\;programmer$ as $woman$ is to $homemaker$" should be reinterpreted in light of this insight.
Previous interpretations took this analogy to be evidence of systematic gender bias in the embedding space \cite{Bolukbasi}. However, there is a very high cosine similarity between $\overrightarrow{man}$ and $\overrightarrow{woman}$ (0.77)\footnote{By comparison, the mean cosine similarity for 100,000 pairs of words randomly sampled from the embedding space was 0.13, with standard deviation 0.11.}; in fact, each is the most similar word to the other in the embedding space. The vectors for $computer\;programmer$ and $homemaker$ are also highly similar (0.50). 
The presence of $homemaker$ can therefore be explained by its similarity to $computer\;programmer$ rather than gender bias. 
Of course, embedding vector similarity does frequently indicate word relatedness (e.g. ``king" and ``queen"). However, vector similarity may also be due to noise. As there is no obvious linguistic relationship between the words $homemaker$ and $computer\;programmer$ and neither are common words in the embedding vocabulary, we posit the latter is the case.

This analogy has been taken as evidence of a gendered relationship between $computer\;programmer$ and $homemaker$ because it has been assumed that the principal relation between the vectors for $man$ and $woman$ is gender, and that this relation carries over to $computer\;programmer$ and $homemaker$. This argument rests on the supposition that the difference vector $\overrightarrow{man} - \overrightarrow{woman}$ encodes gender. However, embeddings were not designed to have such linear properties and their existence has been debated \cite{linzen2016}.  Furthermore, the top solution for ``$man$ is to $apple$ as $woman$  is to ?" is $apples$, but the relationship between $apple$ and $apples$ is clearly pluralisation rather than gender. 
More generally, we took the commonly used Google Analogy Test Set  \cite{Mikolov2013EfficientEO} 
which contains 19,544 analogies (8,869 semantic and 10,675 syntactic) split into 14 categories, such as countries and their capitals.
This set contains 550 unique word  pairs $(x,y)$ (such as $(apple, apples)$) unrelated to gender.\footnote{The category ``family" was excluded as there are gender relationships between the word pairs.} In general, the two words in each of the 550 word pairs are highly similar to each other, with mean cosine similarity 0.62 and standard deviation 0.13. We tested the analogy  ``$man$ is to $x$ as $woman$  is to $?$" using Equation~\ref{eqn:anal}.
 This resulted in 22\% being correctly solved (i.e. returning $y$), including 76\% correct in the ``gram8-plural" category, which contains pluralised words (note that the analogy not being solved correctly does not imply a dissimilar vector is being returned). This demonstrates that ``$man$ is to $x$ as $woman$  is to $?$" frequently solves analogies by returning words whose vectors are similar to $\overrightarrow{x}$, without any need for a linguistically gendered relationship between $x$ and the returned word.

 These observations have further implications for the biased analogy generating method of \newcite{Bolukbasi}, which was extended in \cite{Manzini}. This method leveraged the base pair $(she, he)$ to find word pairs $(x,y)$, such that ``$he$ is to $x$ as $she$ is to $y$", where $||\overrightarrow{x}-\overrightarrow{y}|| = 1$.  However, the condition $||\overrightarrow{x}-\overrightarrow{y}|| = 1$ is equivalent to $\cos(\overrightarrow{x},\overrightarrow{y}) = \frac{1}{2}$. This forced similarity between $x$ and $y$ combined with 
 the high similarity of $she$ and $he$ (0.61) means this method is simply returning word pairs with a high similarity. Alternative choices of gender base pair such as $(woman, man)$ would suffer from the same flaw. Consequently, analogies produced using this method should be treated with caution.
 

	\section{Conclusions}
	There has been a recent focus in the NLP community on identifying  bias in word embeddings. While we strongly support the aim of such work, this paper highlights the complexity of trying to quantify bias in embeddings.
	We showed the reliance of popular gender bias measures on gender base pairs has strong limitations.
None of the measures are robust enough to reliably capture social bias in embeddings,  or to be leveraged in debiasing methods. 
	In addition, we showed the use of gender base pairs to generate ``biased" analogies is flawed.
 Our analysis can contribute to future work designing robust bias measures and effective debiasing methods. 
	Although this paper focused on gender bias, it is relevant to work examining other forms of bias, such as racial stereotyping, in embeddings. Code to replicate our experiments  can be found at: \url{https://github.com/alisonsneyd/Gender\_bias\_word\_embeddings}

\section*{Acknowledgements} This work was supported by the Institute of Coding which received funding from the Office for Students (OfS) in the United Kingdom.

\bibliographystyle{acl_natbib}
\bibliography{aacl-ijcnlp2020}

\begin{thebibliography}{31}
\expandafter\ifx\csname natexlab\endcsname\relax\def\natexlab#1{#1}\fi

\bibitem[{Bem(1974)}]{bem}
Sandra~L. Bem. 1974.
\newblock \href {https://doi.org/10.1037/h0036215} {The measurement of
  psychological androgyny}.
\newblock \emph{Journal of consulting and clinical psychology}, 42:155--62.

\bibitem[{Bolukbasi et~al.(2016)Bolukbasi, Chang, Zou, Saligrama, and
  Kalai}]{Bolukbasi}
Tolga Bolukbasi, Kai-Wei Chang, James Zou, Venkatesh Saligrama, and Adam Kalai.
  2016.
\newblock \href {http://dl.acm.org/citation.cfm?id=3157382.3157584} {Man is to
  computer programmer as woman is to homemaker? debiasing word embeddings}.
\newblock In \emph{Proceedings of the 30th International Conference on Neural
  Information Processing Systems}, NIPS'16, pages 4356--4364, USA. Curran
  Associates Inc.

\bibitem[{Caliskan et~al.(2017)Caliskan, Bryson, and Narayanan}]{Caliskan}
Aylin Caliskan, {Joanna J} Bryson, and Arvind Narayanan. 2017.
\newblock \href {https://doi.org/10.1126/science.aal4230} {Semantics derived
  automatically from language corpora contain human-like biases}.
\newblock \emph{Science}, 356(6334):183--186.

\bibitem[{Cohen(1960)}]{kappa}
Jacob Cohen. 1960.
\newblock Coefficient of agreement for nominal scales.
\newblock \emph{Educational and Psychological Measurement}, 20(1):37–--46.

\bibitem[{Dean and Tate(2016)}]{Dean}
M.~Dean and Charlotte Tate. 2016.
\newblock \href {https://doi.org/10.1007/s11199-016-0713-z} {Extending the
  legacy of sandra bem: Psychological androgyny as a touchstone conceptual
  advance for the study of gender in psychological science}.
\newblock \emph{Sex Roles}, 76.

\bibitem[{Dev and Phillips(2019)}]{attenuatingbis2019}
Sunipa Dev and Jeff~M. Phillips. 2019.
\newblock \href {http://proceedings.mlr.press/v89/dev19a.html} {Attenuating
  bias in word vectors}.
\newblock In \emph{The 22nd International Conference on Artificial Intelligence
  and Statistics, {AISTATS} 2019, 16-18 April 2019, Naha, Okinawa, Japan},
  pages 879--887.

\bibitem[{Devlin et~al.(2019)Devlin, Chang, Lee, and
  Toutanova}]{devlin-etal-2019-bert}
Jacob Devlin, Ming-Wei Chang, Kenton Lee, and Kristina Toutanova. 2019.
\newblock \href {https://doi.org/10.18653/v1/N19-1423} {{BERT}: Pre-training of
  deep bidirectional transformers for language understanding}.
\newblock In \emph{Proceedings of the 2019 Conference of the North {A}merican
  Chapter of the Association for Computational Linguistics: Human Language
  Technologies, Volume 1 (Long and Short Papers)}, pages 4171--4186,
  Minneapolis, Minnesota. Association for Computational Linguistics.

\bibitem[{Drozd et~al.(2016)Drozd, Gladkova, and Matsuoka}]{drozd2016}
Aleksandr Drozd, Anna Gladkova, and Satoshi Matsuoka. 2016.
\newblock \href {https://www.aclweb.org/anthology/C16-1332} {Word embeddings,
  analogies, and machine learning: Beyond king - man + woman = queen}.
\newblock In \emph{Proceedings of {COLING} 2016, the 26th International
  Conference on Computational Linguistics: Technical Papers}, pages 3519--3530,
  Osaka, Japan. The COLING 2016 Organizing Committee.

\bibitem[{Ethayarajh et~al.(2018)Ethayarajh, Duvenaud, and Hirst}]{KawinTU}
Kawin Ethayarajh, David Duvenaud, and Graeme Hirst. 2018.
\newblock \href {http://arxiv.org/abs/1810.04882} {Towards understanding linear
  word analogies}.
\newblock \emph{CoRR}, abs/1810.04882.

\bibitem[{Ethayarajh et~al.(2019)Ethayarajh, Duvenaud, and Hirst}]{Kawin2019}
Kawin Ethayarajh, David Duvenaud, and Graeme Hirst. 2019.
\newblock Understanding undesirable word embedding associations.
\newblock In \emph{Proceedings of the 57th Conference of the Association for
  Computational Linguistics}, pages 1696--1705. Association for Computational
  Linguistics.

\bibitem[{Fleiss(1971)}]{fleiss_kappa}
Joseph~L. Fleiss. 1971.
\newblock {Measuring nominal scale agreement among many raters}.
\newblock \emph{Psychological Bulletin}, 76(5):378--382.

\bibitem[{Gaucher et~al.(2011)Gaucher, Friesen, and Kay}]{Gaucher2011}
Danielle Gaucher, Justin~P Friesen, and Aaron~C. Kay. 2011.
\newblock Evidence that gendered wording in job advertisements exists and
  sustains gender inequality.
\newblock \emph{Journal of personality and social psychology}, 101 1:109--28.

\bibitem[{Gladkova et~al.(2016)Gladkova, Drozd, and Matsuoka}]{gladkova2016}
Anna Gladkova, Aleksandr Drozd, and Satoshi Matsuoka. 2016.
\newblock \href {https://doi.org/10.18653/v1/N16-2002} {Analogy-based detection
  of morphological and semantic relations with word embeddings: what works and
  what doesn{'}t.}
\newblock In \emph{Proceedings of the {NAACL} Student Research Workshop}, pages
  8--15, San Diego, California. Association for Computational Linguistics.

\bibitem[{Gonen and Goldberg(2019)}]{Gonen2019}
Hila Gonen and Yoav Goldberg. 2019.
\newblock Lipstick on a pig: Debiasing methods cover up systematic gender
  biases in word embeddings but do not remove them.
\newblock In \emph{NAACL-HLT}.

\bibitem[{Holt and Ellis(1998)}]{holt1998assessing}
Cheryl~L Holt and Jon~B Ellis. 1998.
\newblock Assessing the current validity of the bem sex-role inventory.
\newblock \emph{Sex roles}, 39(11-12):929--941.

\bibitem[{Kaneko and Bollegala(2019)}]{kaneko2019}
Masahiro Kaneko and Danushka Bollegala. 2019.
\newblock \href {https://doi.org/10.18653/v1/P19-1160} {Gender-preserving
  debiasing for pre-trained word embeddings}.
\newblock In \emph{Proceedings of the 57th Annual Meeting of the Association
  for Computational Linguistics}, pages 1641--1650, Florence, Italy.
  Association for Computational Linguistics.

\bibitem[{Landis and Koch(1977)}]{Landis1977}
J.~Richard Landis and Gary~G. Koch. 1977.
\newblock \href {http://www.jstor.org/stable/2529310} {The measurement of
  observer agreement for categorical data}.
\newblock \emph{Biometrics}, 33(1):159--174.

\bibitem[{Larson and Edwards(2016)}]{larson2016calculus}
R.~Larson and B.H. Edwards. 2016.
\newblock \href {https://books.google.co.uk/books?id=3-S5DQAAQBAJ}
  {\emph{Calculus}}.
\newblock Cengage Learning.

\bibitem[{Levy and Goldberg(2014)}]{levy-goldberg-2014-linguistic}
Omer Levy and Yoav Goldberg. 2014.
\newblock \href {https://doi.org/10.3115/v1/W14-1618} {Linguistic regularities
  in sparse and explicit word representations}.
\newblock In \emph{Proceedings of the Eighteenth Conference on Computational
  Natural Language Learning}, pages 171--180, Ann Arbor, Michigan. Association
  for Computational Linguistics.

\bibitem[{Linzen(2016)}]{linzen2016}
Tal Linzen. 2016.
\newblock \href {https://doi.org/10.18653/v1/W16-2503} {Issues in evaluating
  semantic spaces using word analogies}.
\newblock In \emph{Proceedings of the 1st Workshop on Evaluating Vector-Space
  Representations for {NLP}}, pages 13--18, Berlin, Germany. Association for
  Computational Linguistics.

\bibitem[{Manzini et~al.(2019)Manzini, Lim, Tsvetkov, and Black}]{Manzini}
Thomas Manzini, Yao~Chong Lim, Yulia Tsvetkov, and Alan~W. Black. 2019.
\newblock Black is to criminal as caucasian is to police: Detecting and
  removing multiclass bias in word embeddings.
\newblock In \emph{NAACL-HLT}.

\bibitem[{Matud et~al.(2019)Matud, López-Curbelo, and Fortes}]{Matud_2019}
M.~Pilar Matud, Marisela López-Curbelo, and Demelza Fortes. 2019.
\newblock \href {https://doi.org/10.3390/ijerph16193531} {Gender and
  psychological well-being}.
\newblock \emph{International Journal of Environmental Research and Public
  Health}, 16(19):3531.

\bibitem[{Mehrabi et~al.(2019)Mehrabi, Morstatter, Saxena, Lerman, and
  Galstyan}]{mehrabi2019survey}
Ninareh Mehrabi, Fred Morstatter, Nripsuta Saxena, Kristina Lerman, and Aram
  Galstyan. 2019.
\newblock \href {http://arxiv.org/abs/1908.09635} {A survey on bias and
  fairness in machine learning}.

\bibitem[{Mikolov et~al.(2013{\natexlab{a}})Mikolov, Chen, Corrado, and
  Dean}]{Mikolov2013EfficientEO}
Tomas Mikolov, Kai Chen, Gregory~S. Corrado, and Jeffrey Dean.
  2013{\natexlab{a}}.
\newblock Efficient estimation of word representations in vector space.
\newblock \emph{CoRR}, abs/1301.3781.

\bibitem[{Mikolov et~al.(2013{\natexlab{b}})Mikolov, Sutskever, Chen, Corrado,
  and Dean}]{Mikolov:2013:DRW}
Tomas Mikolov, Ilya Sutskever, Kai Chen, Greg Corrado, and Jeffrey Dean.
  2013{\natexlab{b}}.
\newblock \href {http://dl.acm.org/citation.cfm?id=2999792.2999959}
  {Distributed representations of words and phrases and their
  compositionality}.
\newblock In \emph{Proceedings of the 26th International Conference on Neural
  Information Processing Systems - Volume 2}, NIPS'13, pages 3111--3119, USA.
  Curran Associates Inc.

\bibitem[{Mikolov et~al.(2013{\natexlab{c}})Mikolov, Yih, and
  Zweig}]{mikolov2013linguistic}
Tomas Mikolov, Wen-tau Yih, and Geoffrey Zweig. 2013{\natexlab{c}}.
\newblock Linguistic regularities in continuous space word representations.
\newblock In \emph{Proceedings of the 2013 Conference of the North American
  Chapter of the Association for Computational Linguistics: Human Language
  Technologies}, pages 746--751.

\bibitem[{Nissim et~al.(2019)Nissim, van Noord, and van~der Goot}]{Nissim2019}
Malvina Nissim, Rik van Noord, and Rob van~der Goot. 2019.
\newblock \href {http://arxiv.org/abs/1905.09866} {Fair is better than
  sensational: Man is to doctor as woman is to doctor}.
\newblock \emph{CoRR}, abs/1905.09866.

\bibitem[{Pennington et~al.(2014)Pennington, Socher, and
  Manning}]{pennington2014glove}
Jeffrey Pennington, Richard Socher, and Christopher~D. Manning. 2014.
\newblock \href {http://www.aclweb.org/anthology/D14-1162} {Glove: Global
  vectors for word representation}.
\newblock In \emph{Empirical Methods in Natural Language Processing (EMNLP)},
  pages 1532--1543.

\bibitem[{Peters et~al.(2018)Peters, Neumann, Iyyer, Gardner, Clark, Lee, and
  Zettlemoyer}]{ElMo2018}
Matthew Peters, Mark Neumann, Mohit Iyyer, Matt Gardner, Christopher Clark,
  Kenton Lee, and Luke Zettlemoyer. 2018.
\newblock \href {https://doi.org/10.18653/v1/N18-1202} {Deep contextualized
  word representations}.
\newblock In \emph{Proceedings of the 2018 Conference of the North {A}merican
  Chapter of the Association for Computational Linguistics: Human Language
  Technologies, Volume 1 (Long Papers)}, pages 2227--2237, New Orleans,
  Louisiana. Association for Computational Linguistics.

\bibitem[{Starr and Zurbriggen(2016)}]{Starr}
Christine Starr and Eileen Zurbriggen. 2016.
\newblock \href {https://doi.org/10.1007/s11199-016-0591-4} {Sandra bem’s
  gender schema theory after 34 years: A review of its reach and impact}.
\newblock \emph{Sex Roles}.

\bibitem[{Zhao et~al.(2018)Zhao, Zhou, Li, Wang, and
  Chang}]{zhao2018learningnetrual}
Jieyu Zhao, Yichao Zhou, Zeyu Li, Wei Wang, and Kai-Wei Chang. 2018.
\newblock \href {https://doi.org/10.18653/v1/D18-1521} {Learning gender-neutral
  word embeddings}.
\newblock In \emph{Proceedings of the 2018 Conference on Empirical Methods in
  Natural Language Processing}, pages 4847--4853, Brussels, Belgium.
  Association for Computational Linguistics.

\end{thebibliography}

\appendix
\section{Appendix}
\label{sec:appendix}
\textbf{BSRI Female Terms:} affectionate, affectionately,
cheerful, cheerfully, cheerfulness, childlike,
compassionate, compassionately, feminine, femininely, gentle, gently, gullible, gullibility, gullibly,
loyal, loyally, shy, shyly, shyness, sympathetic,
sympathetically, tender, tenderly, tenderness,
understanding, understandingly, warm, warmish,
warmness, yielding

\vspace{5pt}
\noindent
\textbf{BSRI Male Terms:} aggressive, aggressively,
aggressiveness, aggressivity, ambitious, ambitiously, ambitiousness, analytical, analytically,
assertive, assertiveness, assertively, athletic,
athleticism, athletically, competitive, competitiveness, competitively, dominant, dominantly,
forceful, forcefulness, independent, independently,
individualistic, masculine, selfsufficient

\vspace{5pt}
\noindent
\textbf{Female Animal Terms:} bitch, cow, doe, duck,
ewe, goose, hen, leopardess, lioness, mare, queen,
sow, tigress

\vspace{5pt}
\noindent
\textbf{Male Animal Terms:} dog, bull, buck, drake, ram,
gander, rooster, leopard, lion, stallion, drone, boar,
tiger

\end{document}